\documentclass[letterpaper, 10 pt, conference]{ieeeconf}

\usepackage{graphicx}
\usepackage{amsmath}
\usepackage{amssymb}
\usepackage{cite}
\usepackage[letterpaper,top=54pt,bottom=54pt,left=54pt,right=54pt]{geometry}
\usepackage{epstopdf}
\usepackage[lofdepth,lotdepth]{subfig}
\usepackage{color,soul}
\usepackage{subfig}
\usepackage{lipsum}
\usepackage{wasysym}
\usepackage{hyperref}
\let\labelindent\relax
\usepackage{enumitem}
\usepackage{atbegshi}
\usepackage[font=small,labelfont=bf]{caption}

\AtBeginDocument{\AtBeginShipoutNext{\AtBeginShipoutDiscard}}
\IEEEoverridecommandlockouts
\title{
	\textbf{Tactile-Driven Gentle Grasping for Human-Robot Collaborative Tasks} 
	\vspace{0em}
}

\author{Christopher J. Ford, Haoran Li, John Lloyd, Manuel G. Catalano, Matteo Bianchi,\\ Efi Psomopoulou, Nathan F. Lepora}
\begin{document}

\thanks{CJF was supported by an EPSRC PhD studentship. NL and JL were supported by a Leadership Award from the Leverhulme Trust on ‘A biomimetic forebrain for robot touch’ (RL-2016-39).}
\thanks{CJF, HL, JL, EP and NL are with the Department of Engineering Mathematics and Bristol Robotics Laboratory, University of Bristol, UK (e-mail: chris.ford@bristol.ac.uk, haoran.li@bristol.ac.uk, john.lloyd@bristol.ac.uk, efi.psomopoulou@bristol.ac.uk, n.lepora@bristol.ac.uk). \newline \indent MB is with the Department of Information Engineering and the Research Center "E. Piaggio", University of Pisa, Italy (e-mail: matteo.bianchi@centropiaggio.unipi.it). \newline \indent MGC is with the Department of Soft Robotics for Human Cooperation and Rehabilitation, Istituto Italiano di Tecnologia (IIT), Italy (e-mail: manuel.catalano@iit.it.).}

\maketitle
\thispagestyle{empty}
\pagestyle{empty}
	
\begin{abstract}
This paper presents a control scheme for force sensitive, gentle grasping with a Pisa/IIT anthropomorphic SoftHand equipped with a miniaturised version of the TacTip optical tactile sensor on all five fingertips. The tactile sensors provide high-resolution information about a grasp and how the fingers interact with held objects. We first describe a series of hardware developments for performing asynchronous sensor data acquisition and processing, resulting in a fast control loop sufficient for real-time grasp control. We then develop a novel grasp controller that uses tactile feedback from all five fingertip sensors simultaneously to gently and stably grasp 43 objects of varying geometry and stiffness, which is then applied to a human-to-robot handover task. These developments open the door to more advanced manipulation with underactuated hands via fast reflexive control using high-resolution tactile sensing.
\end{abstract}
\section{INTRODUCTION}

In modern robotics, robust and reliable grasping and manipulation remain unsolved research problems. A central problem is the ability to apply force-sensitive, gentle grasping to handle fragile objects and apply such grasps to objects of various shapes, sizes and stiffnesses \cite{bicchi2000robotic}\cite{chua2003robotic}. Force sensitive grasping is also key in tasks which require humans to work directly with robots. One promising approach to meet these grasping requirements is to use an adaptive soft robotic hand, such as the Pisa/IIT SoftHand \cite{catalano2014adaptive} combined with tactile sensors to provide state information from the contact interface \cite{howe1993tactile}\cite{kappassov2015tactile}. The soft properties of the SoftHand make it ideal for gentle grasping as opposed to other more highly actuated hands with rigid links. 
	
In this paper, we propose that the more limited dexterity of an underactuated hand is compensated by the use of soft synergies that interact with basic control of the hand using fingertip tactile sensing, allowing reliable force-sensitive grasping on a variety of objects. Moreover, human fingertips contain 1000s of tactile mechanoroceptors per square centimetre \cite{wheat1995tactile}, so likewise artificial fingertip tactile sensors carrying similar information should also have a high spatial resolution, as offered by optical tactile sensors such as the TacTip-based design proposed here \cite{ward2018tactip}. Even though there are examples of simple robotic grippers using high-resolution touch for control \cite{donlon2018gelslim}\cite{wilson2020design}, we know of no examples of anthropomorphic hands with multiple high-resolution tactile sensors applied to force-sensitive grasping or the application to a human-robot collaborative tasks. 

This study aims to impart an anthropomorphic robotic hand with reflexive, force-sensitive control to apply a gentle grasp using tactile data from high-resolution sensors. The Pisa/IIT SoftHand used here is under-actuated, yet has soft adaptive synergies in its mechanical structure \cite{catalano2014adaptive} that simplifies the controller implementation whilst retaining a degree of dexterity through its ability to conform to grasped objects. The tactile sensors add an extra capability to this platform by providing information on the nature of the grasp whilst being small and low-cost compared to other works investigating tactile SoftHands \cite{battaglia2017rice}\cite{ajoudani2016reflex}. Here we consider how this can allow for gentle, force-sensitive grasp control on various a priori-unknown objects of differing geometry and stiffness. 

\begin{figure}[t!]
        \centering
        \includegraphics[width = 0.45\textwidth]{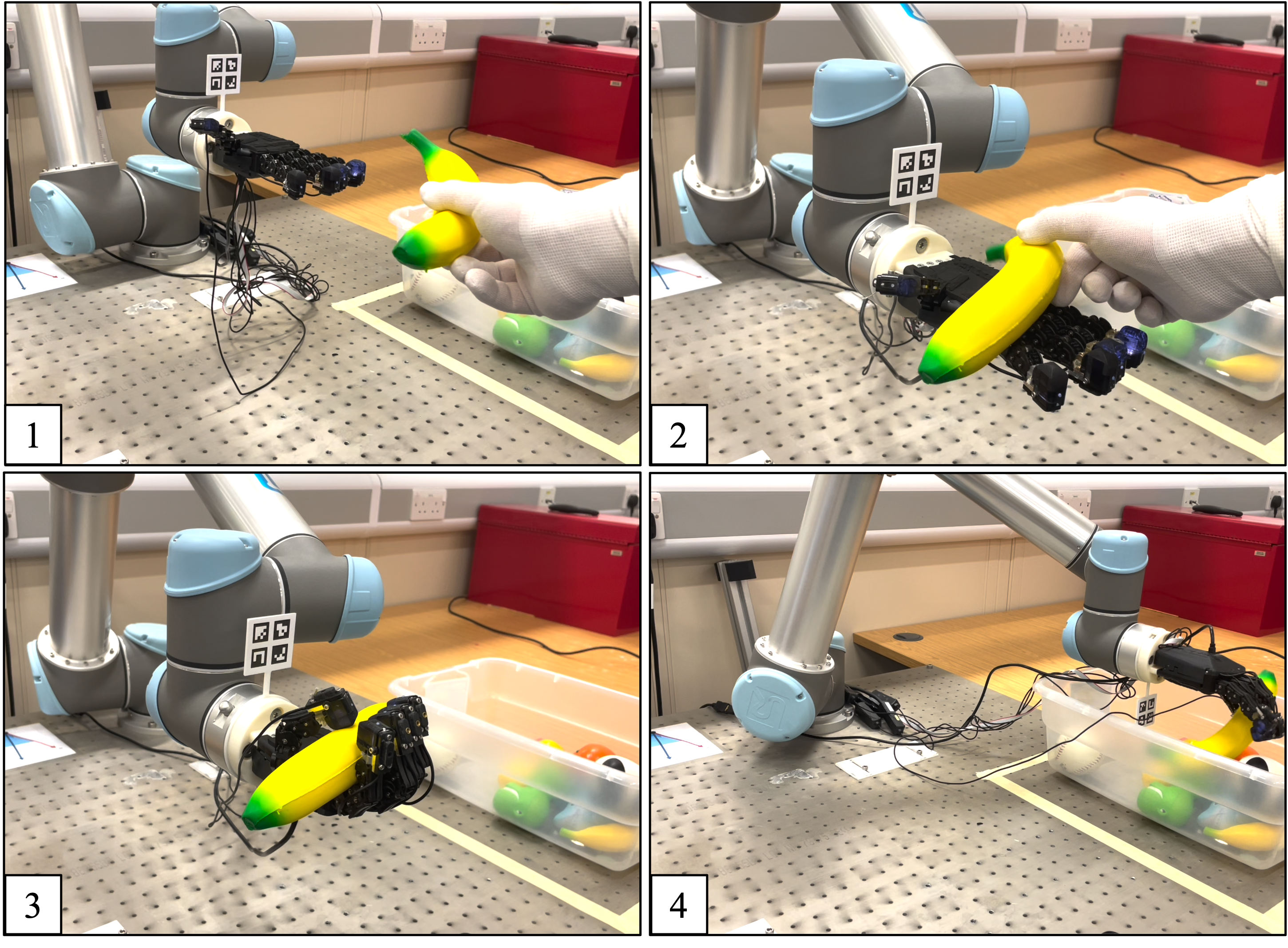}
        \caption{\textit{Tactile-driven handover -} A Pisa/IIT SoftHand with five tactile fingertips performs a handover task with a human operator: (1) safely moving to the operator, (2) accepting, (3) grasping,~then~(4) manipulating a delicate object without applying extraneous force.}
        \label{fig:fig1}
        \vspace{-1em}
\end{figure}
    
Our main contributions are:\\
\noindent \textbf{1) Tactile feedback for force-sensitive grasping:} This work uses tactile feedback from all fingertips to grasp objects without applying extraneous force. This improves upon traditional force-sensitive control methods such as current control by also accounting for the grasp pose, which is important in an underactuated hand.\\
\noindent \textbf{2) Five high-resolution tactile sensors:} Equipping the hand with high-resolution tactile sensors on all fingertips gives detailed feedback on how all digits interact with an object, giving a more-sensitive and capable controller able to understand the nature of the grasp and react accordingly.\\
\noindent \textbf{3) Fast vision-based control loop:} By implementing methods to capture and process tactile data asynchronously whilst also decoupling image capture from the main controller, we propose a system which maximises control loop frequency, resulting in the ability to reliably adjust the grasp quickly and smoothly in response to dynamic changes. 
	
\section{BACKGROUND AND RELATED WORK}
There are many examples of anthropomorphic robotic hands with multi-fingered tactile sensing that use tactile data to perform grasp and subsequent object recognition/classification tasks \cite{zapata2020generation}\cite{jamone2015highly}\cite{devaraja2021design}. However, there are few studies that consider tactile feedback for time-dependent, force sensitive grasp control. Those that do use tactile sensing for grasp control either use sensors providing low-resolution tactile data and/or use a limited number of sensors to mitigate the difficulty of achieving a sufficiently fast control loop for responsive control\cite{santos2019development}\cite{kim2021integrated}\cite{ortenzi2019synergy}\cite{ajoudani2016reflex}.

Human reaction times in response to tactile stimulation are in the range of 150ms-400ms \cite{lele1954reaction}, which sets a benchmark for the performance of robot hands. In this context, Santos and Álvares (2019) \cite{santos2019development} present an anthropomorphic hand with tactile sensing on all five digits that exhibits some manipulation tasks using tactile feedback with a stated response time of 300\,ms (3.3\,Hz). This is within the range of human reaction time described previously and can be considered the present benchmark for tactile reaction time in an anthropomorphic hand with five tactile fingertips. Consequently, the performance of the controller described in this paper will be measured against those criteria.
	
The hand used in this paper is the IIT/Pisa SoftHand: a tendon-driven, underactuated, anthropomorphic soft robotic hand. Designed using principles of postural synergies of human grasping, the SoftHand achieves human-like grasps using one motor that actuates a network of tendons routed through the hand, arranged so that the closure of the hand adheres to a given postural synergy \cite{catalano2014adaptive}. The joints between the phalanges of the fingers are dislocatable, which allows the digits of the hand to conform to grasped objects, creating a soft grasping interface with an ``adaptive" synergy~\cite{della2018toward}. 
	
The sensors used in this paper are a miniaturised version of the BRL TacTip optical tactile sensor, which achieves a high spatial resolution by measuring changes in tactile images over a high-definition pixel array \cite{ward2018tactip,lepora2021soft}. The TacTip is a low-cost, soft, biomimetic tactile sensor that mimics the process of human tactile perception (a camera tracks the movement of an array of pins beneath an artificial skin as an analog to the movement of dermal papillae being relayed as tactile information to the brain via the shallow tactile mechanoreceptors \cite{lepora2021soft}). Previous work from BRL equipped the SoftHand with a single tactile fingertip, feedback from which was used to control grasping through a combined light-touch and pose estimation controller \cite{lepora2021towards} which was successful in proving that tactile feedback from this sensor was capable of accurate grasp control in the context of a generic negative feedback controller. However, there were several limitations, such as a long latency in the control loop, the use of complex deep learning techniques when simpler methods might suffice, and the restriction to just one tactile fingertip that does not permit whole-hand grasp control from contacts on all five digits. 

Here the proposed gentle grasp controller is contextualised and tested in a simplified human-robot handover task. This is a typical human-robot collaborative task and is well suited to a gentle grasp, as a manipulator with a non-force sensitive grasp would be unsafe to use in such close proximity to a human operator. The context of such a task could be in a semi-automated warehouse packing workflow, which may carry additional demands for gentle grasping if the objects being handled are fragile, such as fruits and vegetables. In such environments, humans and robots are often separated for safety reasons, which can be inefficient \cite{huber2008human}.
	
\section{METHODOLOGY}
	
\begin{figure}[t!]
    \centering
    \subfloat[Side view]{{\includegraphics[width=0.2\textwidth]{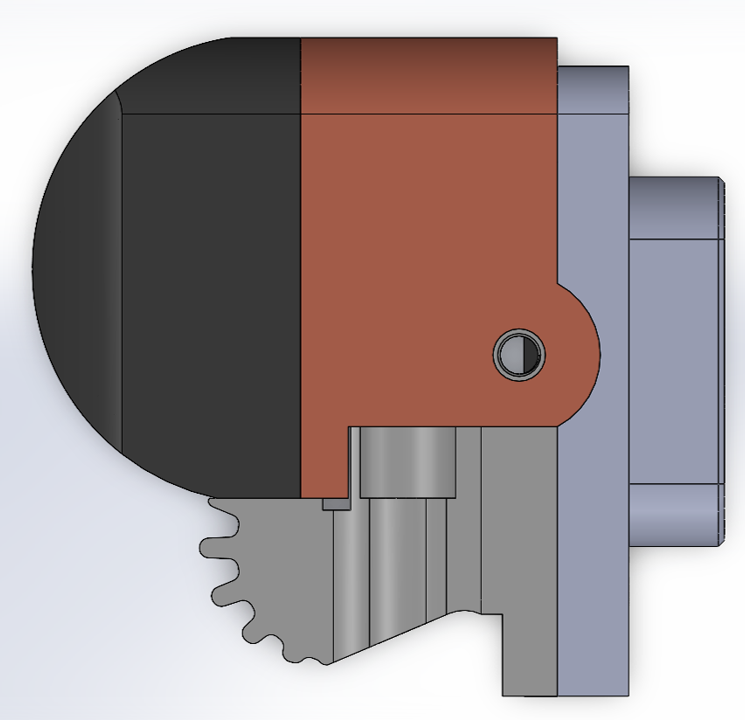} }}
    \qquad
    \subfloat[Side view section]{{\includegraphics[width=0.2\textwidth]{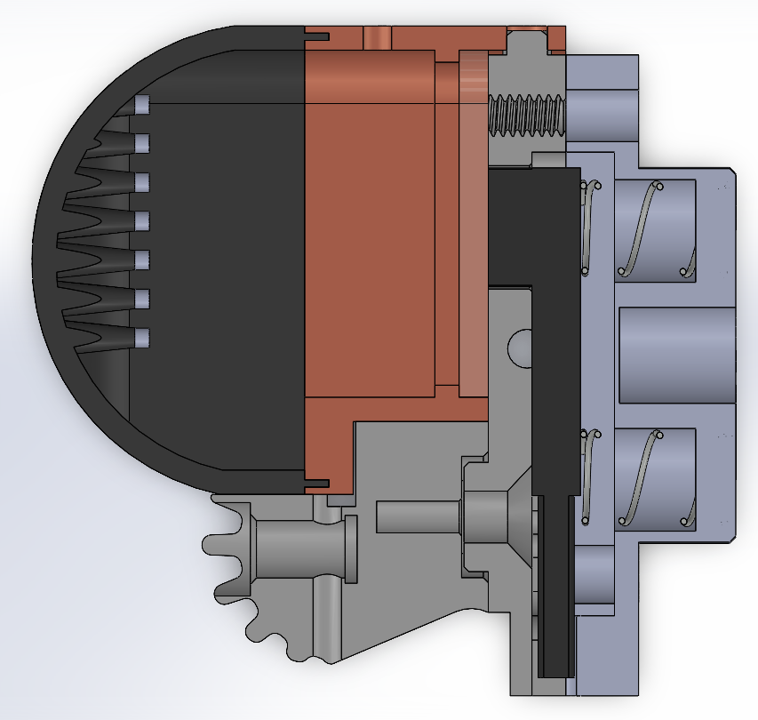} }}
    \qquad
    \subfloat[Exploded view]{{\includegraphics[width=0.45\textwidth]{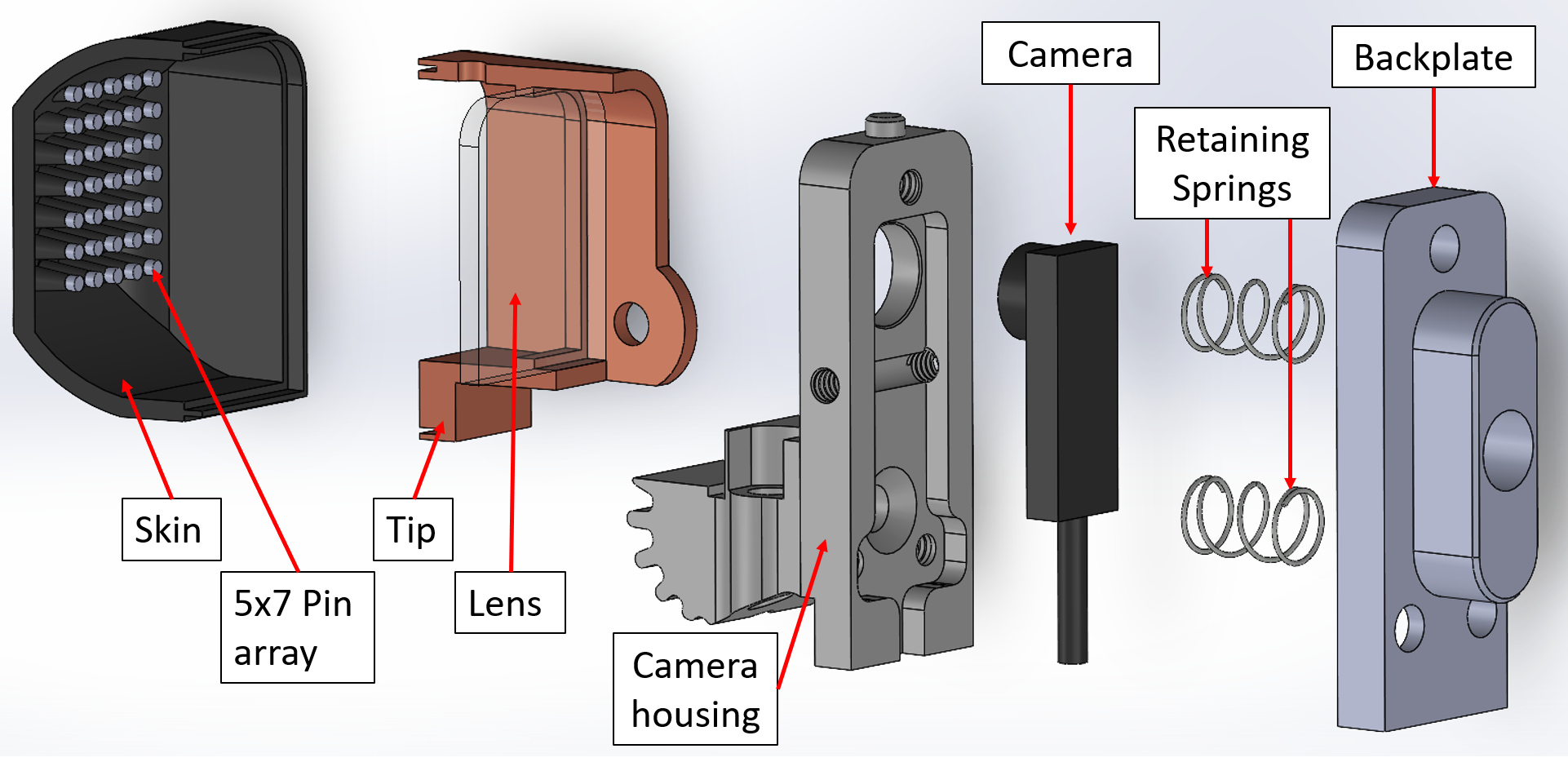} }}
    \caption{\textit{3D model of sensor assembly -} The skin and tip are printed as a single part and filled with an optically-clear gel, through which the camera tracks changes in markers situated on the pin array.}
    \label{fig:cad}
    \vspace{-1em}
\end{figure}
    
\subsection{Tactile Sensors}

The sensors used in this study are an updated iteration of the miniaturised TacTip sensors by Lepora et al. (2021) \cite{lepora2021towards}. Whilst these sensors were successful in proving the capabilities of fingertip tactile sensing on the SoftHand, they were not optimised in mechanical design, such as in the profile of the tactile skin (which was prone to tears due to shearing) and the construction of the sensor as a whole. Consequently, weak points of the design were addressed to give an updated design (shown in Fig. \ref{fig:cad}). The main design changes improved the overall robustness and mechanical stability of the sensor, introducing a new skin material and profile, springs to hold the camera in place and a secure tip-mounting interface. The cameras used in this sensor are Misumi micro-endoscope cameras, which transmit data over USB 3.0 at a native resolution of 1080p.

\begin{figure}[h]
    \centering
    \includegraphics[width = 0.5\textwidth]{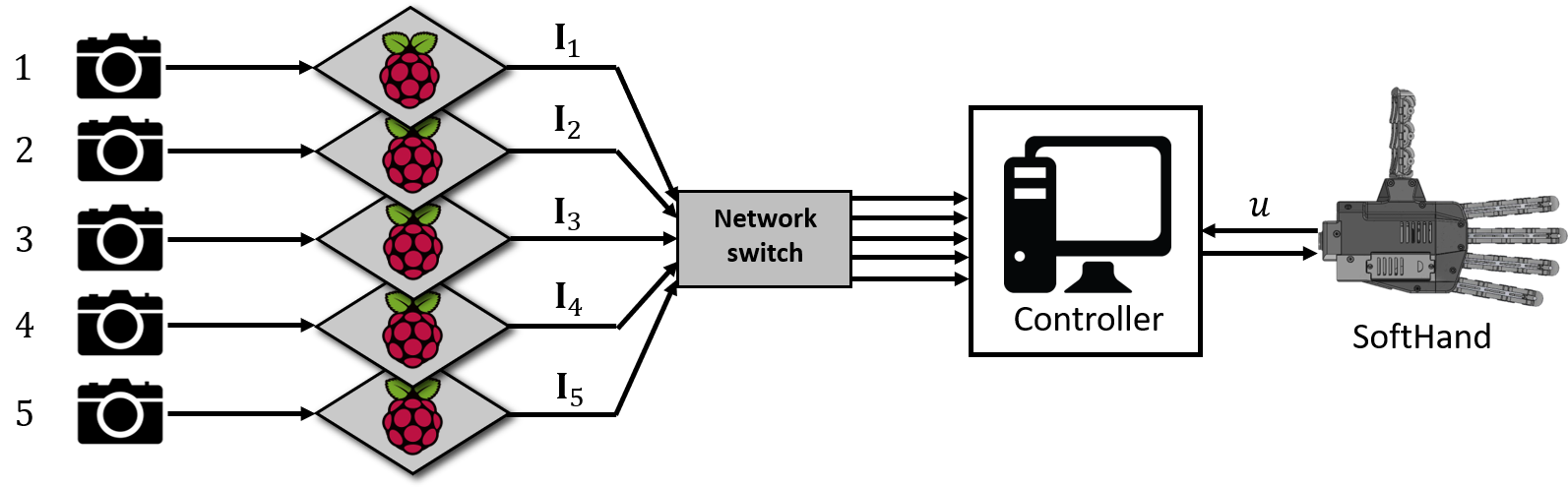}
    \caption{\textit{Asynchronous image capture and processing architecture.} Tactile image data is captured and processed by a dedicated Raspberry Pi per sensor, then sent to the control PC over the network.}
    \label{fig:dataflow}
    \vspace{-1.5em}
\end{figure}
    
\subsection{Data Capture, Transfer and Processing}\label{data}
\vspace{-1em}
Processing multiple optical tactile sensors simultaneously is difficult due to bandwidth restrictions and large processing demands, particularly when using high-resolution tactile images. Consequently, a central problem in this study was to find a way to efficiently capture and process image data from five cameras while using the collective output as feedback into the main control loop, all within a response time that should be as fast as possible.

Preliminary tests showed we could extract and preprocess frames from a single sensor at 30\,fps using our current software; however, if data from each sensor is read sequentially, this process would take over 150\,ms. Whilst this is below the response time used in other studies (see Background), it leaves little time for further processing and would limit the on-line potential of the system. Additionally, by the time the data from the fifth sensor is read, the data taken from the first sensor will be out of sync. Consequently, for best performance, the image data should be extracted from all five cameras asynchronously.

To achieve this, each sensor is connected to its own dedicated Raspberry Pi 4 Model B. The array of Raspberry Pis capture and process image data from all sensors asynchronously, which are then sent to a main control PC via a gigabit Ethernet connection. Using the Raspberry Pi array removed the requirement for the control PC to have a large amount of CPU resource available, as the image capture and processing can be distributed across the processors of the Raspberry Pis, allowing the raw tactile data sampled to be as high quality as possible. By using a distributed computing method (Pyro4) a virtual Python object representing a sensor is registered from each Pi to the network. Then, each sensor object captures and processes tactile images in a background thread and writes the result to a class variable which can be read by the control PC via the network. Whilst the tactile SoftHand's reflex time cannot be faster than a single camera's frame rate (30\,Hz = 33.3\,ms), this method effectively decouples the image capture process from the lower level controller, which prevents the actual control loop time (i.e. the frequency at which the controller output is updated) being limited to the frame rate, resulting in an actual control frequency of 286\,Hz. This is important as it provides scope for expanding the controller with additional operations in the future whilst retaining a real-time response. The architecture of this system is shown in Fig.~\ref{fig:dataflow}.

\begin{figure}[h]
    \centering
    \includegraphics[width = 0.5\textwidth]{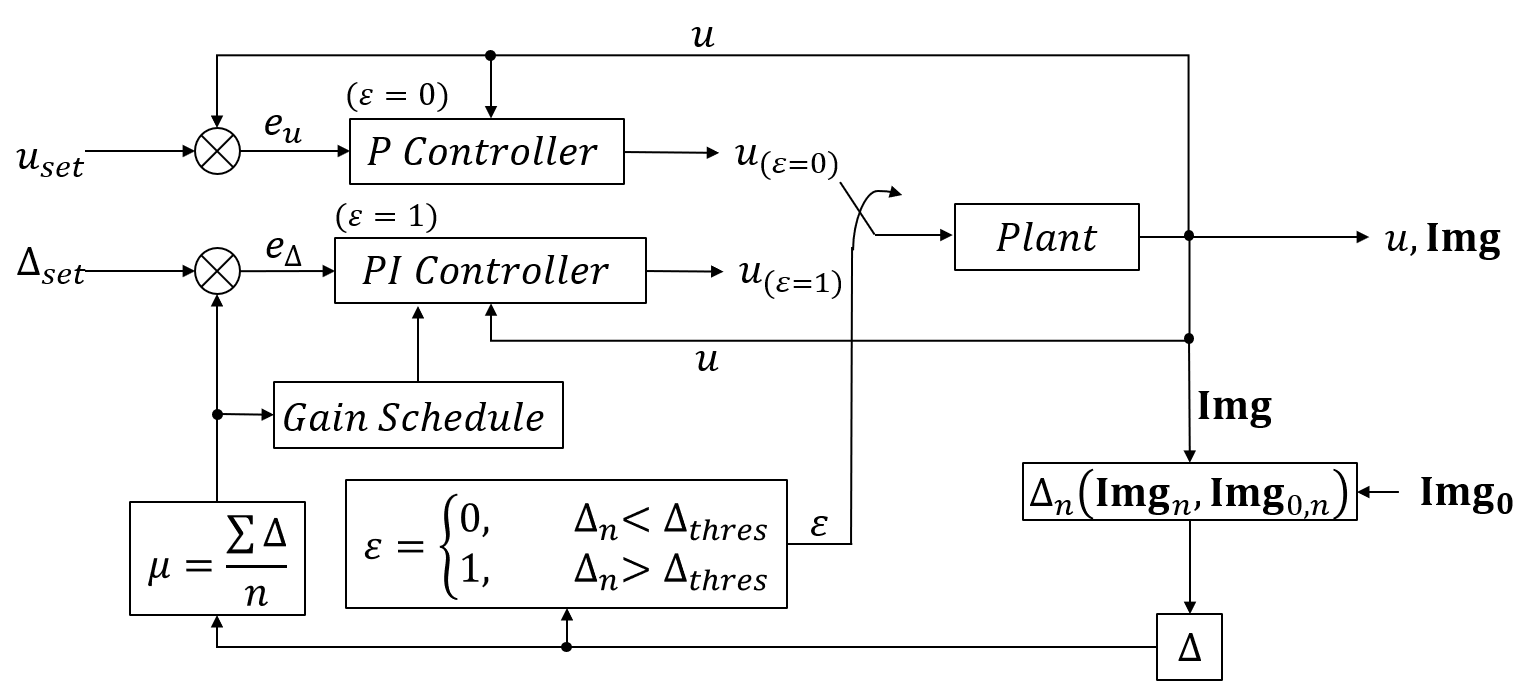}
    \caption{\textit{Controller diagram.} The controller output changes based on a state variable defining if any fingertips are in contact with an object.}
    \label{fig:control}
    \vspace{-1.5em}
\end{figure}
    
\subsection{Controller}\label{controller}

The controller developed in this study uses tactile feedback from all five fingertip sensors simultaneously to apply a stable, gentle grasp on stimuli of various geometries and stiffness. It can also maintain a gentle grasp in response to external disturbances, adapting and adjusting the grasp pose as necessary. 

In previous work with the TacTip on the SoftHand, the main limitation of the controller presented in \cite{lepora2021towards} was a large latency in the system, resulting in a lack of responsiveness. In that previous work, the controller was written in Python, sending and receiving data from the SoftHand via a MEX-compiled Simulink interface. This approach was taken due to the ease of implementation as well as similar applications with the SoftHand \cite{battaglia2017rice}; however, using MATLAB as an intermediary between the main control program and the SoftHand's firmware proved inefficient for reflexive behaviour in this context, resulting in a performance bottleneck.

Hence, the improved controller presented in this work aims for the following capabilities:
\begin{itemize}[leftmargin=*]
    \item[] \textbf{Fast response time:} To address the latency issues seen in previous work, a more direct, low-level programming approach was taken by creating a Python wrapper for a C++ program containing functions from the qbRobotics qbAPI, which communicates with the SoftHand directly. This allows the qbAPI functions to be called directly from the main Python controller, majorly reducing latency.
    \item[] \textbf{Stable response:} By using a state-dependent switching controller, the SoftHand is able to close onto and achieve a gentle grasp on an object quickly and with minimal response overshoot or oscillation.
    \item[] \textbf{Multi-sensor feedback:} By receiving and processing sensor data asynchronously via the methods discussed inSec. IIIB, a control feedback signal is formed from data from all five sensors, enabling the control response to be informed by signals from all five digits.
\end{itemize}
	
The controller functions by measuring changes in tactile images using a Structural Similarity Index Measure (SSIM), an established metric for contact detection in optical tactile sensors \cite{santos2019development} \cite{james2021tactile}. Prior to grasping, one frame from each sensor $n$ is imaged in its undeformed non-contact state $\mathbf{Img_{0,n}}$ and stored for later comparison. As the controller runs, the state of each sensor is imaged $\mathbf{Img_{n}}$ on each time step and compared against its undeformed state, with the SSIM of sensor $n$ against its undeformed state given by: 
\begin{equation}
	    S_n\left( \mathbf{Img_n},\mathbf{Img_{0,n}} \right)=\frac{\left( 2\mu_x\mu_y+c_1 \right)\left( 2\sigma_{xy}+c_2 \right)}{(\mathrm{\mu}_{x}^{2}+\mathrm{\mu}_{y}^{2}+c_1)(\mathrm{\sigma}_{x}^{2}+\mathrm{\sigma}_{y}^{2}+c_2)}
	    \label{eq:ssim}
\end{equation}
where $x$ and $y$ represent a square kernel of pixels (here of size 7$\times$7) that is applied across both images as a sliding window. The parameters $\sigma$ and $\mu$ represent the mean and covariance of each kernel calculation, with $c_1$ and $c_2$ acting as regularising constants to stabilise the division \cite{james2021tactile}. The SSIM is given as an averaged final value, $S_n \in [0,1]$, where $S_n = 1$ indicates the two images are identical and $S_n = 0$ indicates they have zero similarity \cite{wang2004image}. 

Here the SSIM-based metric used to control the hand is derived from a modified version of Equation (\ref{eq:ssim}) to represent the degree of deformation, denoted as $\Delta_n$ and given by:
\begin{equation}
    \Delta_n\left( \mathbf{Img_n},\mathbf{Img_{0,n}} \right) = 1-S_n\left( \mathbf{Img_n},\mathbf{Img_{0,n}} \right).
\end{equation}
The output is inverted from the SSIM so that a value closer to 1 indicates a greater amount of sensor deformation and thus a larger contact force. At each time step, a $5\times1$ vector containing $\Delta_n$ for all five sensors is calculated asynchronously to minimise the control loop time (Fig.~ \ref{fig:control}, bottom right).
	
In a grasping task, each finger can be in one of two states: in contact or not in contact with an object. When the fingers are not in contact with an object (state 0), the system should behave so as to move the fingers into a contacted state (state 1) as quickly as possible. Once state 1 has been achieved, fast yet fine motor movements should be made to achieve and maintain a stable grasp. Since these two states have differing dynamic requirements, we use a state-dependent switching controller (see Fig.~\ref{fig:control}) to streamline the process and  achieve the desired system behaviour. 

This controller design is computationally similar to a finite state machine: switching between two linear controllers according to state variable $\varepsilon$ that signifies whether any fingertip contacts an object. When $\varepsilon=0$, {\em i.e.} no contact is detected, the hand is driven using a proportional controller using the motor encoder position, $u$, as feedback with a defined setpoint of maximum closure. When $\varepsilon=1$, {\em i.e.} one or more sensors are in contact with an object, the controller switches to a PI controller that uses the average degree of deformation,  $\mu =\sum_n\Delta_n/{n}$ as feedback. The potential behavioural issues that are sometimes associated with this type of controller are addressed by constantly keeping track of feedback variables for both states regardless of the occupied control mode, eliminating discontinuities in the control signal \cite{leith2003issues}. Based on previous work \cite{james2021tactile}, the threshold condition for a sensor $n$ being in contact with an object was defined as $\Delta_n>0.05$. Additionally, a value $\mu = 0.5$ was chosen as a setpoint for the entire controller, as preliminary tests suggested this value would be applicable to applying a gentle grasp across a wide range of objects.

\subsection{Handover task}
\begin{figure}[h]
    \centering
    \includegraphics[width = 0.45\textwidth]{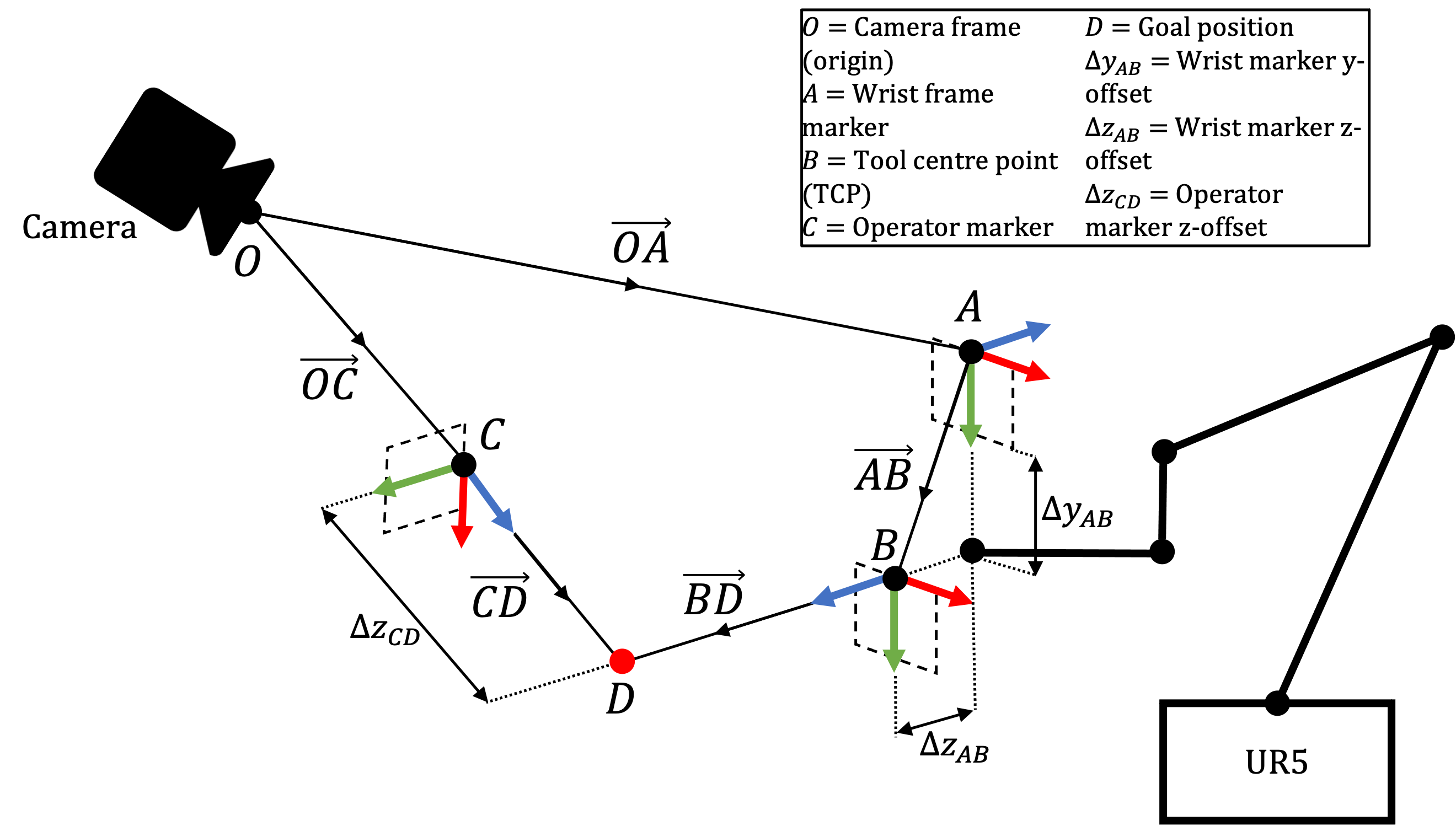}
    \caption{\textit{Experiment 2 setup.} In order to move the SoftHand to the operator, the 3D transformation between the two must be solved.}
    \label{fig:ex2setup}
    \vspace{-1em}
\end{figure}

\begin{figure*}[t!]
    \centering
    \includegraphics[width = \textwidth]{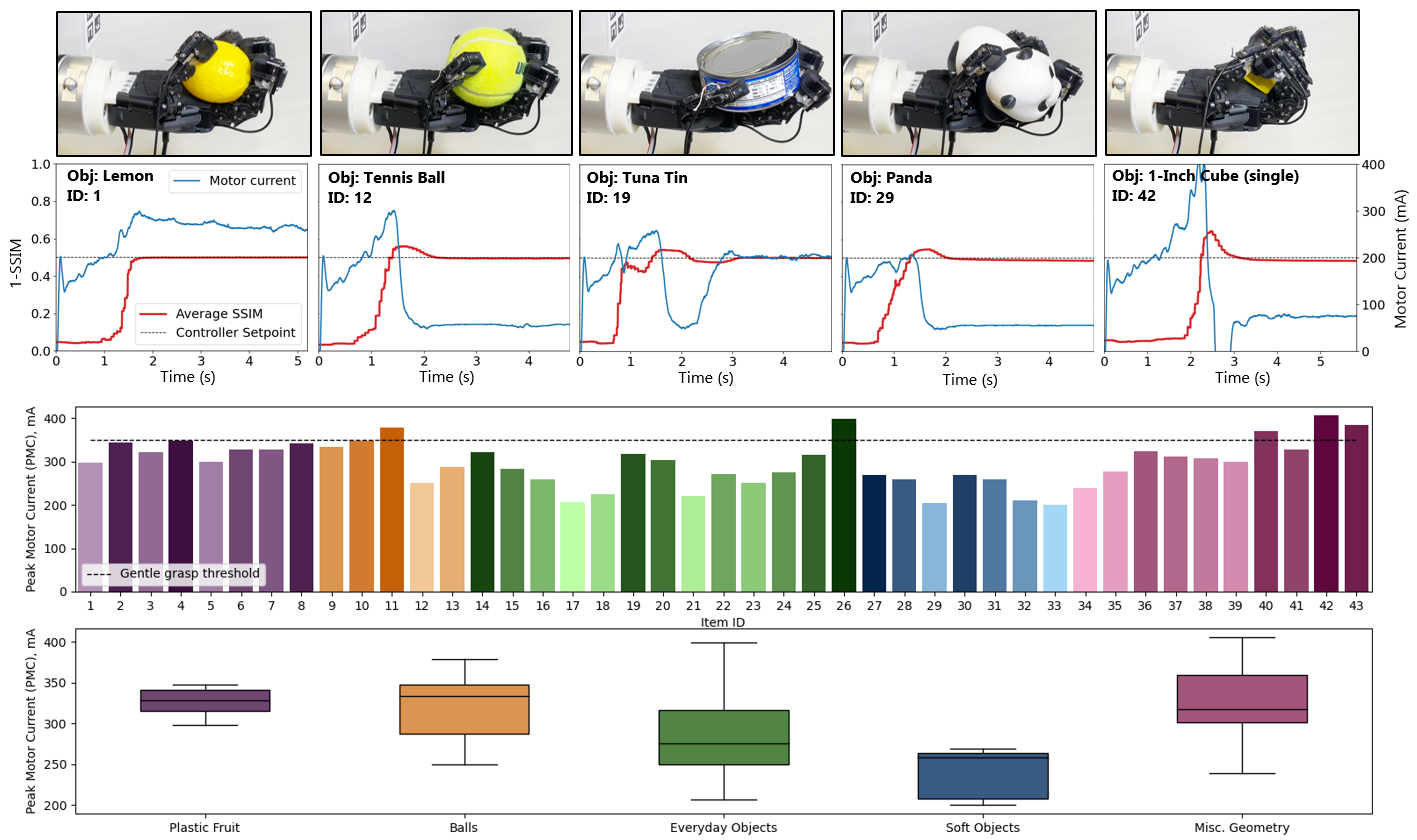}
    \caption{\textit{Experiment 1 Results. }The top row of plots show typical control responses on an object from each category. The bar chart shows the average PMC of each object used in this experiment (which represents peak grasping force) with an associated box-and-whisker plot to illustrate the spread of values for each category.}
    \label{fig:mainResult}
    \vspace{-1em}
\end{figure*}

The handover task used to test the validated gentle grasp controller consists of the SoftHand mounted to a UR5 industrial robot arm (ideally suited to this purpose as it is a `cobot', meaning it has safety features which allow it to interface directly with human operators \cite{ur5}) which will move to the hand of a human operator, accept an object from them, then carry and deposit the object in a bin. The task itself was kept simple, as the focus was intended to test the capability of the grasp controller, with the handover acting as a contextual scenario.

To track the position of the SoftHand relative to the operator, two ArUco markers were used: one offset yet co-planar to the robot's wrist frame and another attached to the back of a glove worn on the hand of the operator that holds the object, with both tracked by an Intel RealSense RGBD camera (3D geometric diagram of the setup shown in Fig.~\ref{fig:ex2setup}).

In terms of the robot's movement, only translational motion was considered when moving to the operator due to inconsistencies in detecting relative rotational pose using just ArUco marker data, amendments to which lie outside the scope of this work. In order to move to the goal position, the vector $\overrightarrow{BD}$ must be solved. $\overrightarrow{BD}$ is calculated by extracting the pose of the markers $A$ and $C$ relative to the camera frame $O$ and calculating the corresponding homogeneous transformation matrices, $T_{OA}$ and $T_{OC}$, the translational components of which correspond to $\overrightarrow{OA}$ and $\overrightarrow{OC}$ respectively. Due to the planar constraints between $B$ and $D$ relative to $A$ and $C$ respectively, the vectors $\overrightarrow{AB}$ and $\overrightarrow{CD}$ can be found from the homogenous transformation matrices which describe the linear transformation between the points $(A,B)$ and $(C,D)$, calculated by:
\begin{eqnarray}
    &T_{AB}=T_{OA}\cdot\begin{bmatrix}
 \mathrm{I}_{3}& \mathrm{\begin{bmatrix}
0 &\Delta y_{AB}  &\Delta z_{AB} 
\end{bmatrix}}^{T} \\
 \begin{matrix}
0 &0  &0 
\end{matrix}&1 
\end{bmatrix},\\
    &T_{CD}=T_{OC}\cdot\begin{bmatrix}
 \mathrm{I}_{3}& \mathrm{\begin{bmatrix}
0 &0  &\Delta z_{CD} 
\end{bmatrix}}^{T} \\
 \begin{matrix}
0 &0  &0 
\end{matrix}&1 
\end{bmatrix}.
\end{eqnarray}
Then $\overrightarrow{AB}$ and $\overrightarrow{CD}$ are extracted as the translational components of $T_{AB}$ and $T_{CD}$ respectively, with $\overrightarrow{BD}$ found by
\begin{equation}
    \overrightarrow{BD} = \overrightarrow{CD} - \overrightarrow{AB}.
\end{equation}

Since the position of $B$ relative to the base frame of the robot is known through the robot's kinematics, the robot can be programmed to execute a Cartesian movement to the goal position by giving it the co-ordinates of $\overrightarrow{BD}$.

The robot arrives at the goal position in a palm-up orientation, at which point the operator can place the object into the hand, which then grasps the object using the gentle grasp controller established in section \ref{controller}. Once the grasp is stable, the arm executes a Cartesian movement to the bin where it releases the grasp to deposit the object. Since the SoftHand is palm-up when accepting the object, this movement involves rotating the hand 180 degrees to be palm-down when depositing. During this motion, the grasp controller continues to run and adjust the grasp as necessary.

\section{RESULTS}
\subsection{Experiment 1: Stable, gentle grasping via tactile control}

The first experiment sought to assess the validity of the controller; namely, its ability to apply a stable, gentle grasp on a variety of stimuli of differing geometry and stiffness. The 43 objects used in this task are listed in Fig.~\ref{fig:ex2results}, where each are associated with one of five categories and denominated with a numerical ID. These objects were selected to assess controller performance on different sizes, topologies and stiffnesses. The wide range of objects was important to test, as a linear controller output in the SoftHand does not correlate to a linear grasp-pose profile. In each trial, the object was placed in the most natural position within the grasping envelope in approximately the same orientation for each trial, with five trials performed for each object.
    
Fig.~ \ref{fig:mainResult} shows typical examples of the control responses for objects from each category. In each example, the controller reaches stability from first contact within 1-3\,s (defined as the response remaining between a threshold of $\pm5\%$ of the setpoint after initial contact).
    
Fig.~\ref{fig:mainResult} also shows the average Peak Motor Currents (PMC) for each object. Previous work with the SoftHand has focused on force-sensitive grasp control using the current draw of the motor to estimate grasping force \cite{mura2018soft} \cite{della2017estimating}. Whilst motor current does give a representation of grasp force (with PMC correlating to peak force), our tests show that the underactuated nature of the hand causes this measurement to not be invariant of the contact distribution of the fingers. This results in an aliasing effect, whereby a given motor current can be associated with a variety of grasp poses and contact distributions dependent of the orientation of the object with respect to the hand. Using tactile feedback avoids this issue, as it allows a gentle grasp to be applied whilst also providing information on how the individual fingers are interacting with the object, giving a more reliable picture of grasp pose without the need for additional sensors. 

Considering the argument for a gentle grasp controller driven by current feedback, the results of this experiment show that the PMC required for a gentle grasp can vary by up to 200\,mA depending on the size, shape and stiffness of the object. The box and whisker plot in Fig.~\ref{fig:mainResult} further emphasises this, showing that PMC is lower for soft objects (aligning with observations in \cite{mura2018soft}) and that the spread of PMC is greater for object sets with a more diverse range of sizes and stiffnesses.  Additionally, the tactile feedback controller consistently results in a PMC below the 350\,mA `gentle grasp' threshold for the SoftHand defined by \cite{mura2018soft} despite motor current being absent from the feedback loop. The controller sometimes exhibited higher PMC (for small objects) and oscillations while stabilising, such as with Items 13 and 34. The greater closure with smaller objects required higher peak motor currents, which we attribute to the higher torque demands from the greater extension of the elastic links when the hand is near closed. Overall, this shows that the SSIM deformation vector $\mu$ is valid as a feedback signal for a tactile-driven gentle grasp controller.
	
\subsection{Experiment 2: Tactile-driven handover task}
\begin{figure}[t!]
    \centering
    \includegraphics[width = 0.45\textwidth]{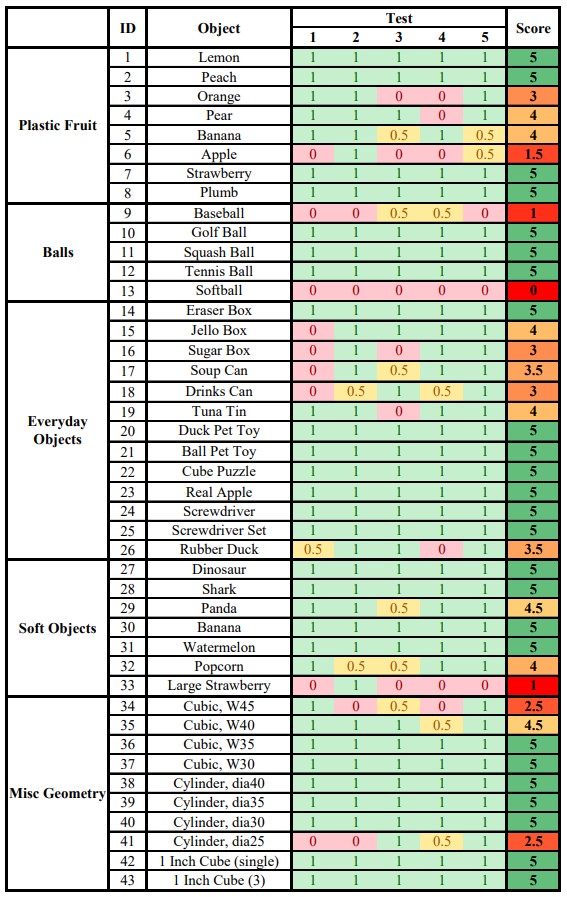}
    \caption{\textit{Experiment 2 Results.} Each item from the object set was tested five times in the handover task, with performance measured by cumulative pass/fail criteria.}
    \label{fig:ex2results}
    \vspace{-1em}
\end{figure}

The handover task for Experiment 2 was performed with each object used in Experiment 1 with five trials per object. For each trial, the success of the system was scored by awarding 1 point if the object was successfully deposited in the bin, 0 points for a failure ({\em i.e.} if the object slips out of the grasp and does not end up in the bin or if the controller fails to stabilise) and 0.5 points for a partial success (exhibited by the object slipping from the grasp but still ending up in the bin). This gives a total score out of 5, which describes how well the system performed in handling the object. The results for Experiment 2 are shown in Fig.~\ref{fig:ex2results}.

The results show good performance across most of the tested objects, with many being successful on all trials. Overall, the system was successful in the handover task on 80\% of trials when counting absolute successes only and 87\% when also counting partial successes. The failures seemed to occur when handling objects that were large in comparison to the SoftHand's grasp envelope (Item 11) or objects of higher mass (Item 9), as the force exerted by the object on the fingertips was interpreted as overloading and caused the grasp to release early. This is inherent to the controller design, yet the effect is reduced by using the average SSIM of all sensors as feedback, meaning failures only occur on objects of significant mass. The partial successes seemed to occur on objects that were large but also on longer objects which tended to slip out of the grasp. However, as shown in the results the controller performed well overall and these edge cases could be addressed with further work. 
    	
\section{DISCUSSION}
In this paper, we presented a novel method for achieving stable, force-sensitive grasping using tactile feedback for an underactuated, anthropomorphic soft robotic hand. The controller achieves and maintains a stable, gentle grasp on objects, adjusting grasp pose as necessary to adapt to dynamic changes without applying extraneous forces, achieved without the complex tuning and methods associated with more traditional current-control approaches. This was implemented with an effective control loop time of 286\,Hz, which is far below the minimum control loop of 3.3\,Hz seen in other studies \cite{santos2019development}, which helps achieve the dynamic, reflexive behaviour in response to changing grasp conditions.
	
A key observation of this study was how using tactile feedback significantly decreased the complexity of the controller application for gentle grasping. In all experiments, the motor current rarely exceeded the gentle grasp threshold of 350\,mA defined in \cite{mura2018soft} despite motor current being absent from the control loop. The controller was capable of consistently applying a gentle grasp on 43 distinctly different objects (and even on a cluster of 3 small objects, in Item 43). The interaction between the measure of deformation of the tactile sensors and the semi-coupled nature of the SoftHand's digits through an adaptive synergy combined with the controller is capable of applying force-sensitive grasps to a range of objects without the need for monitoring current. 
	
There were some cases where the control methods could be improved with further work, namely the high PMCs seen on small objects and the failed handover results. These could be remedied by accounting for degree of hand closure and implementing a slip detection driven controller for dynamic movements, as described by James et. al (2018)\cite{james2018slip}\cite{james2020slip}.
	
Another direction for extending this work would be a more sophisticated handover task. This could include the robot physically taking the object from the operator or considering SoftHand orientation relative to the operator in the approach, as seen in \cite{sidiropoulos2019human} and \cite{psomopoulou2015human}. Overall, the methods presented in this paper open the door to more advanced manipulation with underactuated anthropomorphic hands through feedback from high-resolution tactile sensors on each fingertip, taking another step towards human-like dexterous manipulation with robots.
\vfill	
	
\bibliographystyle{IEEEtran}
\bibliography{IEEEabrv,ref}
\end{document}